\documentclass{article}
\usepackage{graphicx} 
\usepackage{url}

\usepackage{listings}
\usepackage{xcolor}
\usepackage{graphicx}

\usepackage{geometry}                
\usepackage{amssymb}
\usepackage{epstopdf}
\usepackage{soul}
\usepackage{amsmath,amsbsy}
\usepackage[shortlabels]{enumitem}
\usepackage{multirow}
\usepackage{longtable}
\usepackage{subfigure}
\usepackage{amsmath}

\usepackage{makecell}

\lstdefinestyle{mystyle}{
    language=Python,
    numbers=left,
    numberstyle=\small,
    numbersep=8pt,
    showspaces=false,
    showstringspaces=false,
    breaklines=true,
    frame=single,
    basicstyle=\ttfamily\small,
    keywordstyle=\color{blue}\bfseries,
    stringstyle=\color{red},
    commentstyle=\color{green},
}

\begin{document}

\title{Balanced Multimodal Learning via Mutual Information}
\author{Rongrong Xie and Guido Sanguinetti\\
\textit{Scuola Internazionale Superiore di Studi Avanzati (SISSA)}
}
\date{}

\maketitle

\begin{abstract}%
Multimodal learning has increasingly become a focal point in research, primarily due to its ability to integrate complementary information from diverse modalities. Nevertheless, modality imbalance, stemming from factors such as insufficient data acquisition and disparities in data quality, has often been inadequately addressed. This issue is particularly prominent in biological data analysis, where datasets are frequently limited, costly to acquire, and inherently heterogeneous in quality. Conventional multimodal methodologies typically fall short in concurrently harnessing intermodal synergies and effectively resolving modality conflicts. 

In this study, we propose a novel unified framework explicitly designed to address modality imbalance by utilizing mutual information to quantify interactions between modalities. Our approach adopts a balanced multimodal learning strategy comprising two key stages: cross-modal knowledge distillation (KD) and a multitask-like training paradigm. During the cross-modal KD pretraining phase, stronger modalities are leveraged to enhance the predictive capabilities of weaker modalities. Subsequently, our primary training phase employs a multitask-like learning mechanism, dynamically calibrating gradient contributions based on modality-specific performance metrics and intermodal mutual information. This approach effectively alleviates modality imbalance, thereby significantly improving overall multimodal model performance. 
\end{abstract}

\section{Introduction}

Multimodal learning aims to integrate complementary signals from diverse data types, yet in practice one modality often dominates training when information content, data quality, or sample size are imbalanced. This \emph{modality imbalance} suppresses the benefits of integration and is especially problematic in biomedical applications such as multi-omics disease subtyping, where cohorts are small and assays vary in noise and coverage. Foundational syntheses emphasize fusion, alignment, and coordination as core challenges, but principled mechanisms that explicitly counter modality imbalance while preserving useful cross-modal structure remain limited \cite{baltruvsaitis2018multimodal}.

We propose a balanced multimodal framework for multi-omics classification that combines three ideas: (i) graph-based encoders that exploit cross-sample structure; (ii) cross-modal knowledge transfer to strengthen weaker modalities; and (iii) a multitask-style optimization procedure that adaptively reweights unimodal and multimodal losses based on performance signals and cross-modal dependence. Concretely, we employ a revised graph convolutional encoder in which node features may derive from a single modality, while edges are constructed from a fused similarity network across modalities. We then pretrain weaker modalities via \emph{knowledge distillation} from a stronger teacher to transfer predictive structure without overfitting \cite{hinton2015distilling,furlanello2018born}. Finally, we train the joint model with dynamic loss balancing so that no single modality dictates the gradients, leveraging advances in multitask optimization \cite{chen2018gradnorm,kendall2018multi}.

A central component of our approach is quantifying and utilizing cross-modal statistical dependence. We use mutual information (MI) both to guide when knowledge transfer is plausible and to modulate training so that fusion emphasizes informative relations. Modern MI estimators (e.g., MINE) and InfoNCE-style contrastive objectives make such dependence measurable and tractable in deep models, and MI maximization has improved multimodal fusion in prior work \cite{belghazi2018mutual,oord2018representation,han2021improving}.

We evaluate our method on breast cancer (BRCA) classification using copy-number variation (CNV), mRNA expression, and proteomic (RPPA) profiles assembled from community resources that promote reproducibility (UCSC Xena and TCPA) \cite{goldman2020visualizing,li2017explore}. Across experiments, the combination of fused-graph encoders, MI-aware distillation, and adaptive reweighting mitigates modality imbalance and improves macro-F1 relative to naive early/late fusion and unimodal baselines, while remaining robust under low-information channels and small-sample regimes common in oncology.

More broadly, our results suggest a practical recipe for multimodal learning under scarcity and heterogeneity: \emph{transfer what is reliable}, \emph{quantify what is shared}, and \emph{balance what is optimized}. Although we focus on bulk multi-omics, the same principles extend to other settings (e.g., single-cell multi-omics, imaging--genomics, or clinical multimodal records) and can be coupled with probabilistic integration frameworks (e.g., MOFA+) to accommodate missingness and uncertainty \cite{argelaguet2020mofa+}.

\section{Method}

\subsection{Overall Framework and Problem Formulation}

Integrating multi-omics data for classification tasks is challenging due to the inherent imbalance and varying levels of informativeness among different data modalities. Consider a dataset with $N$ samples, each represented across $M$ modalities denoted as $X=\{X^1, X^2, ..., X^m, ..., X^M\}$. All modalities share the same ground truth label $y$ among $C$ distinct classes. Our goal is to develop a robust classification framework that effectively leverages all modalities while addressing the imbalance problem.

\begin{figure}
\centering
\includegraphics[width=0.95\textwidth,angle=0]{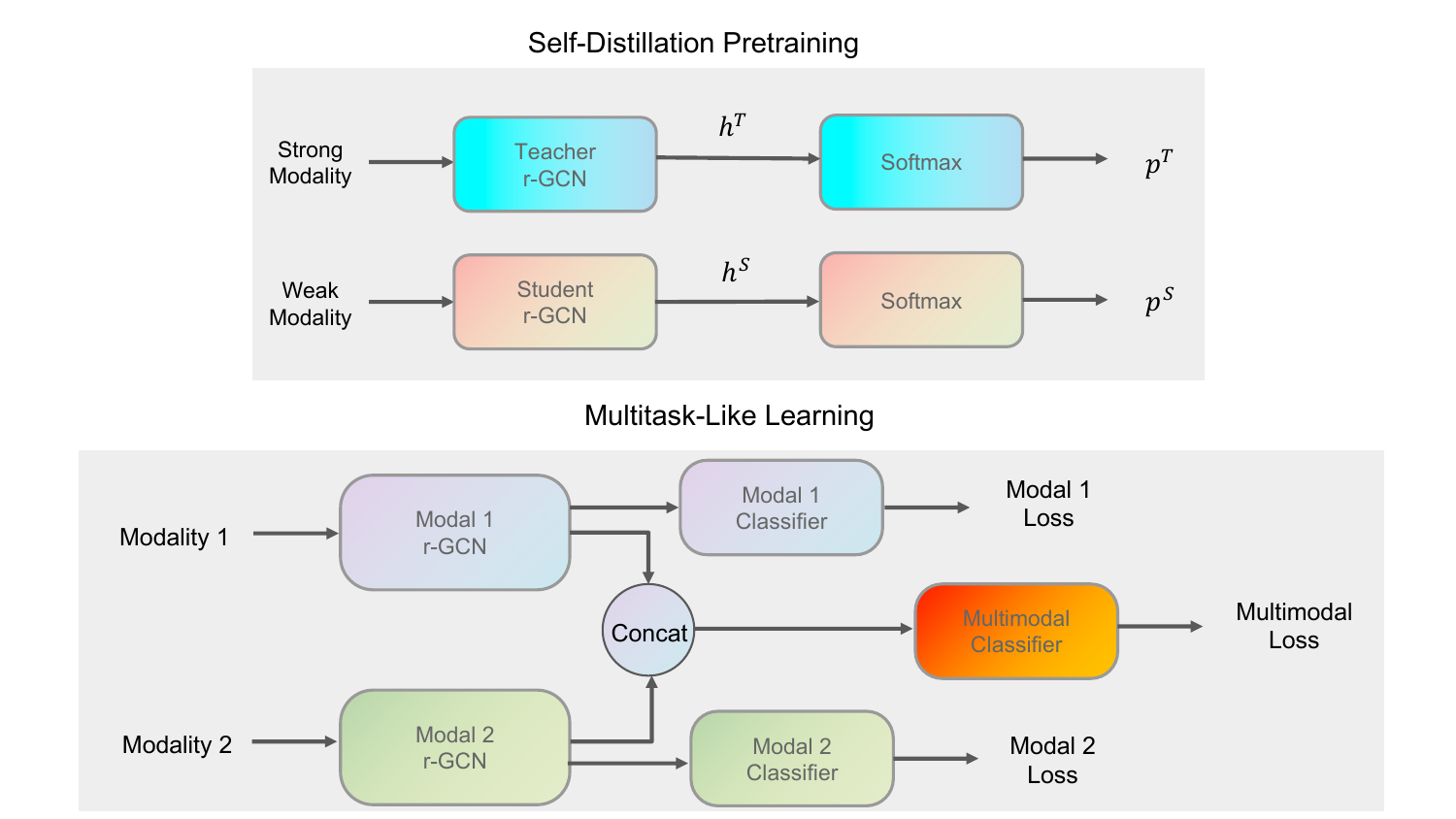}
\caption{Framework.}
\label{fig:framework}
\end{figure}

To achieve this, we propose a unified framework that incorporates a revised Graph Convolutional Network (r-GCN) as the unimodal encoder for each modality (illustrated in Figure \ref{fig:framework}). Detailed in Section \ref{subsec:r_GCN}, this r-GCN enhances feature representations by integrating multi-modal similarities into its graph structure, enabling each modality to capture more informative and discriminative patterns. Alongside the r-GCN, our framework addresses modality imbalance through two key strategies, elaborated in Section \ref{subsec:balanced_multimodal}: (1) Self-Distillation Pretraining, where weaker modalities learn from stronger ones by distilling knowledge from their pretrained models, and (2) a Multitask-Like Learning Framework that dynamically adjusts the learning contributions of each modality based on performance metrics such as the macro F1 score.

\subsection{revised Graph Convolution Network}
\label{subsec:r_GCN}
\paragraph{Similarity Network Fusion}

Similarity Network Fusion \cite{wang2014similarity} (SNF) is a computational method used to integrate multiple modalities into a unified network, which is particularly useful in biomedical applications for understanding complex biological systems.

When fusing two modalities, SNF begins by constructing a similarity matrix $\mathbf{W}$ for each modality, quantifying the similarity between samples:
\begin{equation}
\mathbf{W}(i, j) = \exp\left(-\dfrac{\rho^2(x_i, x_j)}{\mu \, \epsilon_{i,j}}\right),
\label{exp_matrix}
\end{equation}
where $\rho(x_i, x_j)$ is the Euclidean distance between samples $x_i$ and $x_j$, $\mu$ is a hyperparameter, and $\epsilon_{i,j}$ adjusts for scaling effects.

A normalized matrix $\mathbf{P}$ is then defined:
\begin{equation*}
\mathbf{P}(i, j) =
\begin{cases}
\dfrac{\mathbf{W}(i, j)}{2 \sum\limits_{k \neq i} \mathbf{W}(i, k)}, & \text{if } j \neq i \\[2ex]
\dfrac{1}{2}, & \text{if } j = i
\end{cases}
\end{equation*}

To emphasize local similarities, a $K$-nearest neighbors matrix $\mathbf{S}$ is computed:
\begin{equation*}
\mathbf{S}(i, j) =
\begin{cases}
\dfrac{\mathbf{W}(i, j)}{\sum\limits_{k \in N_i} \mathbf{W}(i, k)}, & \text{if } j \in N_i \\[2ex]
0, & \text{otherwise}
\end{cases}
\end{equation*}
where $N_i$ contains the $K$ nearest neighbors of sample $i$.

The fusion process iteratively updates the similarity matrices:
\begin{equation*}
\begin{aligned}
\mathbf{P}_{t+1}^{1} &= \mathbf{S}^{1} \, \mathbf{P}_t^{2} \left( \mathbf{S}^{1} \right)^\top, \\
\mathbf{P}_{t+1}^{2} &= \mathbf{S}^{2} \, \mathbf{P}_t^{1} \left( \mathbf{S}^{2} \right)^\top.
\end{aligned}
\end{equation*}
Here, $\mathbf{P}_t^{1}$ and $\mathbf{P}_t^{2}$ represent the status matrices for the first and second modalities, respectively, at iteration $t$.

After convergence, the integrated network is obtained by averaging:
\begin{equation}
\mathbf{P}^{c} = \dfrac{\mathbf{P}_t^{1} + \mathbf{P}_t^{2}}{2}.
\label{state_matrix}
\end{equation}
This results in a unified similarity matrix that effectively combines multiple modalities, enhancing the representation of relationships among samples.

\paragraph{Graph Convolutional Networks}
Graph Convolutional Networks \cite{kipf2016semi} (GCNs) have become a fundamental framework for analyzing graph-structured data, allowing for the extraction of salient patterns and representations through the simultaneous utilization of node features and the graph topology. The training of a GCN model requires two essential inputs: the feature matrix \(X \in \mathbb{R}^{N \times B}\), where \(N\) denotes the number of nodes and \(B\) represents the dimensionality of the feature space associated with each node, and the adjacency matrix \(A \in \mathbb{R}^{N \times N}\), which encapsulates the graph's structural information.

The GCN architecture typically consists of multiple convolutional layers. The propagation rule for the \(l\)-th layer is expressed as:

\[
H^{(l+1)} = \sigma \left( \tilde{D}^{-\frac{1}{2}} (\tilde{A}) \tilde{D}^{-\frac{1}{2}} H^{(l)} W^{(l)} \right),
\]

where \(\tilde{A} = A + I_N\) is the adjacency matrix with added self-loops, \(I_N\) is the identity matrix of size \(N\), \(\tilde{D}\) represents the corresponding degree matrix, and \(W^{(l)}\) is the trainable weight matrix at layer \(l\). The function \(\sigma\) is the non-linear activation function, typically the Rectified Linear Unit (ReLU).

In this formulation, the input to each layer \(H^{(l)}\) is iteratively updated, with the initial input \(H^{(0)}\) set to the feature matrix \(X\). This propagation mechanism enables GCNs to effectively integrate and transform node features across the graph, capturing both local neighborhood information and broader structural patterns.

For datasets lacking a predefined adjacency matrix, the adjacency element \(A_{ij}\) between nodes \(i\) and \(j\) can be computed as follows:

\[
A_{ij} =
\begin{cases} 
s(\mathbf{x}_i, \mathbf{x}_j), & \text{if } i \neq j \text{ and } s(\mathbf{x}_i, \mathbf{x}_j) \geq \epsilon, \\
0, & \text{otherwise},
\end{cases}
\]

where \(\mathbf{x}_i\) and \(\mathbf{x}_j\) are the feature vectors corresponding to nodes \(i\) and \(j\), respectively. For unimodal, the similarity network \(s(\mathbf{x}_i, \mathbf{x}_j)\) can be computed using the scaled exponential similarity, as given in Eq.(\ref{exp_matrix}). For multimodal, the similarity can be derived via the SNF algorithm, detailed in Eq.(\ref{state_matrix}). The threshold \(\epsilon\) is typically determined based on the desired average number of edges per node \cite{wang2021mogonet}.

\paragraph{revised Graph Convolution Network}

As previously discussed,  GCNs utilize node features and the relationships between nodes, often represented by a similarity matrix, to perform various tasks. However, in the context of biological data, such as transcriptomic data, we have observed that the similarity matrix often exhibits low values, which fails to effectively capture the relationships between nodes.

In response to this limitation, we propose a revised GCN (r-GCN) approach wherein the strict relationship between nodes and edges is relaxed. In this revised model, node features may originate from a single modality, while edge weights are derived from a similarity matrix that integrates multiple modalities. This integration is achieved using the SNF algorithm, thereby enhancing the representation of relationships in the network.

\subsection{Balanced Multimodal Learning} 
\label{subsec:balanced_multimodal}

Imbalanced multimodal learning remains a significant challenge due to the varying levels of informativeness and performance across different modalities. Current methods typically address this issue by either controlling the training of unimodal encoders based on inter-modal performance discrepancies or by reducing the impact of low-informative modalities. In this paper, we present a novel and comprehensive methodology that integrates both aspects. Initially, we utilize the macro F1 score to assess the learning state of each modality, categorizing them into strong, weak, or low-information groups. We then employ self-distillation to pre-train the modalities, allowing weaker modalities to benefit from the knowledge of stronger ones. Finally, we apply a multitask-like multimodal framework that dynamically adjusts the gradient magnitudes for each modality, ensuring a balanced and effective multimodal learning process.

\subsubsection{Estimating the Modality Learning State}

To evaluate the learning state of each modality, we employ the \emph{macro F1 score} due to its robustness and straightforward application.

Previous approaches have often estimated unimodal learning states by analyzing differences between training and validation datasets. For example, \cite{wang2020makes} utilized the \emph{overfitting-to-generalization ratio}, calculated from the training loss \(L^{Tr}\) and validation loss \(L^{Va}\) at epochs \(e\) and \(e+n\):
\[
O_e = L_e^{Va} - L_e^{Tr}, \quad \text{Ratio} = \left| \frac{O_{e+n} - O_e}{L_e^{Va} - L_{e+n}^{Va}} \right|.
\]
Similarly, \cite{wei2024diagnosing} proposed the \emph{purity gap} \(g^m\), defined as the difference between the training purity \(P_{Tr}^m\) and validation purity \(P_{Va}^m\), where purity measures how well the representations cluster according to the true class labels.

However, in our experiments with omics data, we found that these metrics did not effectively reflect the true learning states of different modalities. Specifically, the overfitting-to-generalization ratio and purity gap yielded similar values across modalities, despite variations in their actual performance. This indicates that these metrics are insufficiently sensitive to the nuances of each modality's learning process, limiting their utility in our context.

In contrast, the macro F1 score overcomes these limitations by providing a straightforward and accurate measure of each modality's true performance. Furthermore, unlike the softmax-based prediction scores of each modality \cite{peng2022balanced}, it offers a balanced evaluation even in datasets with class imbalances. Therefore, we advocate for the use of the macro F1 score as a more reliable and effective metric for estimating unimodal learning states in multimodal learning settings with imbalanced datasets.

\subsubsection{Multimodal Pre-training with Self-Distillation}

To address the challenge of imbalanced multimodal learning, we employ a self-distillation strategy \cite{liu2023multimodal} to transfer knowledge from a strong modality to a weaker one during pre-training. This approach leverages the rich information captured by the strong modality to enhance the learning of the weaker modality.

We designate the modality with the highest macro F1 score as the \emph{strong modality}. If a modality's macro F1 score is close to that of random guessing, i.e., $1/C$ where $C$ is the number of classes, it indicates that it contains very little useful information; we refer to it as a \emph{low-information modality}. Modalities with macro F1 scores between the strong modality and the low-information modality are termed \emph{weak modalities}. In this work, we apply self-distillation to improve the performance of weak modalities with the assistance of the strong modality. The details are illustrated in Figure~\ref{fig:framework}.

We first pre-train a teacher model using the strong modality. For each sample $i$ in the strong modality, the teacher model extracts a representation $h_i^T$. The predictions for the $C$ classes are obtained by applying the SoftMax function to $h_i^T$, i.e.,
\[
p_i^{T} = \frac{e^{h_i^{T}}}{\sum_{j=1}^{C} e^{h_j^{T}}}.
\]
After pre-training the teacher model, we train a student model using the weak modality. During training, the loss function $L$ is defined as:
\begin{align*}
L &= \alpha_1 L_{CE} + \alpha_2 L_{KL} + \alpha_3 L_{RE}, \\
L_{CE} &= -\sum_{i=1}^{N} y_i^S \log p_i^S, \\
L_{KL} &= \sum_{i=1}^{N} D_{KL}(p_i^T \| p_i^S), \\
L_{RE} &= \sum_{i=1}^{N} \frac{\| h_i^T - h_i^S \|^2}{d}.
\end{align*}
Here, $L_{CE}$ computes the cross-entropy loss for the student model, where $y_i^S$ is the true label for sample $i$ and $p_i^S$ is the predicted probability from the student model. The terms $L_{KL}$ and $L_{RE}$ are the self-distillation components, where the student model attempts to mimic the knowledge (prediction distributions and representations) of the teacher model. Specifically, $L_{KL}$ is the Kullback–Leibler divergence between the teacher's predicted probabilities $p_i^T$ and the student's predicted probabilities $p_i^S$. The term $L_{RE}$ is the Euclidean distance between the representations of the teacher model $h_i^T$ and the student model $h_i^S$, normalized by the dimension $d$ of the representations. The weighting factors $\alpha_1$, $\alpha_2$, and $\alpha_3$ are used to balance the contributions of each loss component.

We refrain from applying self-distillation strategies to low-information modalities because they have limited capacity and low correlation with the strong modality. However, if a low-information modality exhibits sufficient correlation with the strong modality—such as a mutual information greater than 0.2—it can still benefit from the self-distillation strategy.

\subsubsection{Multitask-like Multimodal Framework}
After utilizing self-distillation for pre-training the modalities, we employ a multitask-like multimodal framework \cite{wang2020makes} to further address the issue of imbalanced multimodal learning. In the multitask-like framework illustrated in Figure~\ref{fig:framework}, each modality $\mathbf{X}^m$ is fed into its corresponding encoder $\varphi^m(\mathbf{X}^m)$ parameterized by $\boldsymbol{\theta}^m$ to extract features. These extracted features are then fused using a fusion operation $\oplus$ (e.g., concatenation) and passed to a multimodal classifier $f^{M+1}(\varphi^{1} \oplus \varphi^{2} \oplus \dots \oplus \varphi^{M})$, where $M+1$ denotes the multimodal classifier.

Simultaneously, the features from each modality's encoder $\varphi^m(\mathbf{X}^m)$ are also passed to their respective unimodal classifiers $f^{m}(\varphi^{m}(\mathbf{X}^{m}))$. In this multitask-like multimodal framework, we consider not only the multimodal loss $L^{M+1} = L(f^{M+1}(\varphi^{1} \oplus \varphi^{2} \oplus \dots \oplus \varphi^{M}), y)$ but also the optimization of each unimodal classifier, incorporating the unimodal losses $L^m = L(f^{m}(\varphi^{m}(\mathbf{X}^m)), y)$ for $m = 1, \dots, M+1$. Therefore, the total loss of the model is:

\begin{equation*}
L = \sum_{m=1}^{M+1} k_t^m L^m,
\end{equation*}
where $k_t^m$ is a coefficient used to reweight each loss term. By reweighting the losses, we can control the magnitude of gradients from each task (both multimodal and unimodal), thereby reducing gradient conflicts. This balancing mechanism leads to more synergistic updates and has been empirically verified to effectively alleviate the imbalanced multimodal learning problem \cite{wei2024mmpareto}.

Regarding the definition of the coefficient $k_t^m$, previous works often focus either on dominant and weak modalities \cite{peng2022balanced,kontras2024improving} or on low-information modalities \cite{wei2024diagnosing}. In contrast, we propose a novel and more comprehensive approach that combines both considerations. We define $k_t^m$ as follows:
\begin{align*}
r_t^m &= \frac{F^m}{\frac{1}{M-1} \sum_{j=1, j \neq m}^{M} F^j}, \\
k_t^m &= \begin{cases}
1 - \tanh(\alpha \cdot r_t^m), & \text{if } F^m > \gamma \frac{1}{C}, \\
\tanh(\beta \cdot r_t^m), & \text{otherwise},
\end{cases}
\end{align*}
where $F^m$ is the macro F1 score of modality $m$, $r_t^m$ is the ratio of the macro F1 score of modality $m$ compared to the average macro F1 score of the other modalities, and $\tanh$ provides a smooth normalization of $r_t^m$. The parameters $\alpha$, $\beta$, and $\gamma$ are hyperparameters.

By defining $k_t^m$ in this manner, we ensure that low-information modalities receive a lower weighting (by setting a small $\beta$), thereby decreasing their contribution to the overall loss and preventing them from adversely affecting the model's performance. Additionally, this approach decreases the weight of strong modalities and increases the weight of weak modalities, mitigating the dominance of strong modalities in the optimization process. Consequently, weaker modalities receive sufficient optimization efforts, allowing them to gain adequate training.

\section{Experiment}

\subsection{Dataset and implementation details}
In this study, we utilize the BRCA dataset, accessible via the UCSC Xena browser (https://xenabrowser.net/) and the Cancer Proteome Atlas (TCPA) portal (https://tcpaportal.org/tcpa/). This dataset, focused on breast tumors, encompasses four distinct molecular subtypes: (1) Basal-like, characterized by the absence of hormone receptor and ERBB2 expression; (2) HER2-enriched, notable for its overexpression of the oncogene ERBB2; (3) Luminal A; and (4) Luminal B. Both Luminal A and B subtypes are generally estrogen receptor (ER)-positive and express epithelial markers, although Luminal B exhibits a higher Ki67 index and a less favorable prognosis compared to Luminal A. For our analysis, we selected multi-omic data across three biological levels: genomic (copy number variation, CNV), transcriptomic (mRNA sequencing), and proteomic (reverse phase protein array, RPPA). A total of 511 patients with data available across all three omics were included in the study. Detailed information regarding the dataset is provided in Table \ref{tab_numSamples}.

\begin{table}
  \centering \caption{The number of samples for each subtype of BRCA.}\label{tab_numSamples}
  \begin{tabular}{lccccc}
\hline
  & Basal-like  & Her2-enriched & Luminal A & Luminal B & Total \\ \hline
 Number of samples  & 112 & 53 & 248 & 98 & 511 \\ \hline
\end{tabular}
\end{table}

\subsection{Baseline}
The dimensional characteristics of each modality within the BRCA dataset are presented in Table \ref{tab_numFeatures}. Given the high dimensionality of this data, substantial computational resources are required for analysis. To mitigate the computational burden, we employed an autoencoder to reduce the dimensionality of each modality to 100 features. Recognizing that different modalities may exhibit varying levels of classification performance, we utilized a random forest classifier as a benchmark to evaluate and compare the classification performance of individual modalities as well as various combinations of these modalities, see the accuracy, roc-auc, and average f1 results in Table \ref{tab_rf}.

Based on Table \ref{tab_rf}, it is evident that the classification performance of the CNV modality is significantly lower compared to mRNA and RPPA across all metrics, including accuracy, AUC, and Macro F1. When CNV is combined with other modalities, the performance improvements are minimal or, in some cases, even result in a decrease compared to combinations that exclude CNV. This suggests a clear imbalance among the modalities. The CNV modality appears to introduce a substantial amount of noise, which adversely affects the model's predictive capability, leading to poorer overall performance. Therefore, if the noise inherent in the CNV modality is not appropriately addressed, and all modalities are combined indiscriminately for classification tasks, the expected improvement in classification performance may not be realized, and it may, in fact, deteriorate.

\begin{table}
  \centering \caption{The number of features for each modality of BRCA.}\label{tab_numFeatures}
  \begin{tabular}{lcccc}
\hline
& mRNA & CNV & RPPA & \\ \hline
Number of features & 19580 & 19273 & 223 & \\ \hline
\end{tabular}
\end{table}

\begin{table}
  \centering \caption{Classification performance across various modalities using the logistic regression.}\label{tab_rf}
  \begin{tabular}{lccc}
\hline
& Accuracy & AUC & Macro F1 \\ \hline
mRNA & $ 0.8338 \pm 0.0340 $ & $0.9510\pm0.0115 $ & $0.7476\pm0.0504 $ \\ \hline
CNV & $0.5831\pm0.0338 $ & $0.6763\pm0.0221 $ & $0.3206\pm0.0168 $\\ \hline
RPPA & $0.8434\pm0.0231 $ & $0.9602\pm0.0067 $ & $0.8130\pm0.0282 $\\ \hline
mRNA+CNV & $0.8338\pm0.0354 $ & $0.9503\pm0.0090 $ & $0.7576\pm0.0412 $\\ \hline
RPPA+CNV & $0.8493\pm0.0203 $ & $0.9603\pm0.0065 $ & $0.8233\pm0.0201 $\\ \hline
RPPA+mRNA & $0.8728\pm0.0156 $ & $0.9737\pm0.0050 $ & $0.8403\pm0.0249 $\\ \hline
RPPA+mRNA+CNV & $0.8728\pm0.0174 $ & $0.9735\pm0.0053 $ & $0.8367\pm0.0292 $\\ \hline
\end{tabular}
\end{table}

\subsection{revised Graph convolution network}
We view each sample as a node within a graph, wherein both the node features and the inter-nodal relationships—established via a similarity network—are input into a GCN to perform classification tasks. As illustrated in Table \ref{tab_GCN}, two distinct types of node features are considered: single modality and combined modalities. For the single modality, we employ a scaled exponential similarity algorithm to construct a weighted similarity network. In contrast, for the combined modalities, we utilize the SNF algorithm to generate a fusion-weighted similarity network.

\begin{figure}
\centering
\includegraphics[width=0.95\textwidth,angle=0]{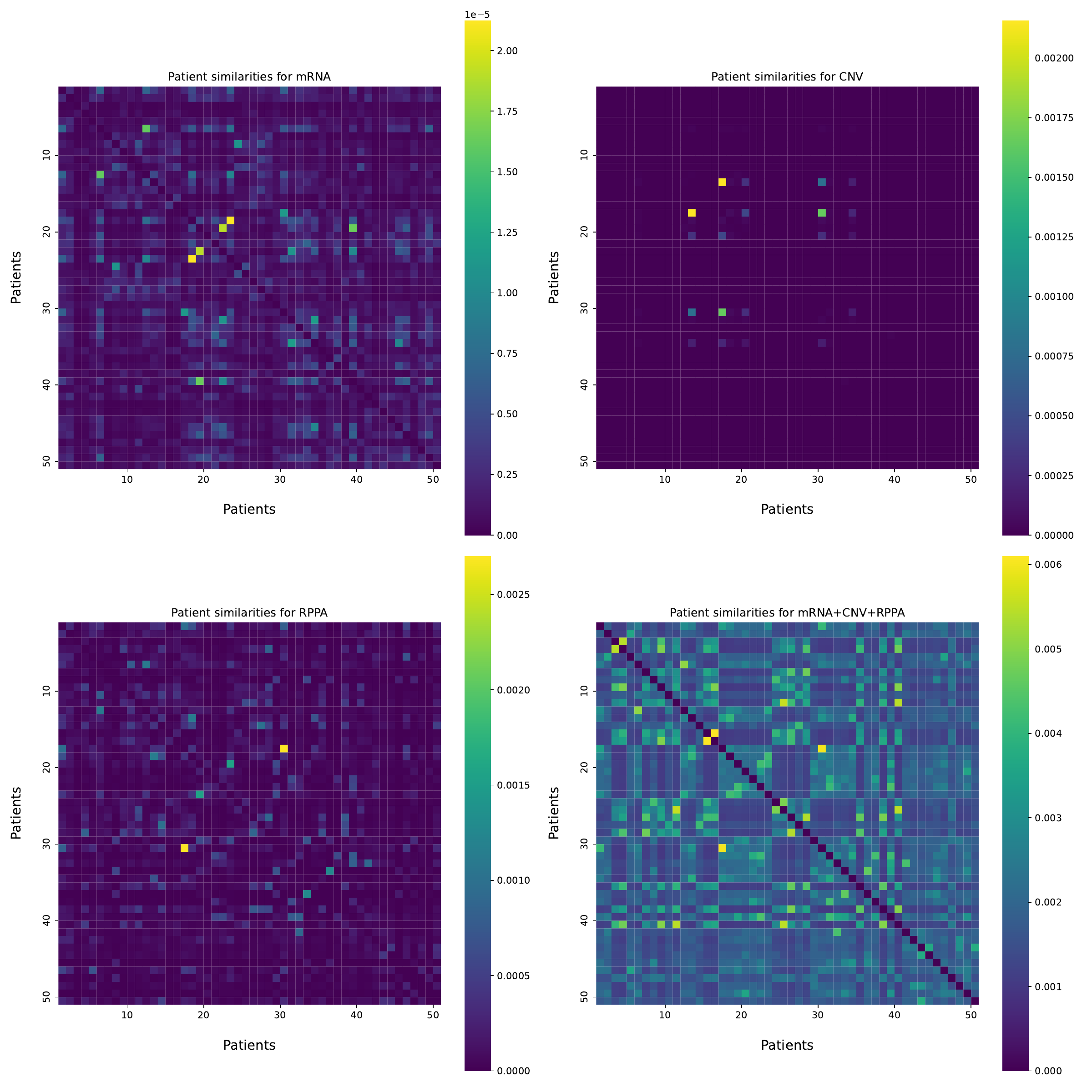}
\caption{Heatmap comparison across various modalities.}
\label{fig_heatmap}
\end{figure}

The results presented in Table \ref{tab_GCN} indicate that the GCN outperforms the random forest algorithm in classification accuracy. Nonetheless, as depicted in Figure \ref{fig_heatmap}, the patient similarities within the single modality similarity network exhibit weaker associations (i.e., low-weight edges) when compared to those observed in the fusion-weighted similarity network generated via SNF. This discrepancy raises our concerns regarding the effectiveness of the single modality similarity network in accurately representing the relationships between samples.

\begin{figure}
\centering
\includegraphics[width=0.95\textwidth,angle=0]{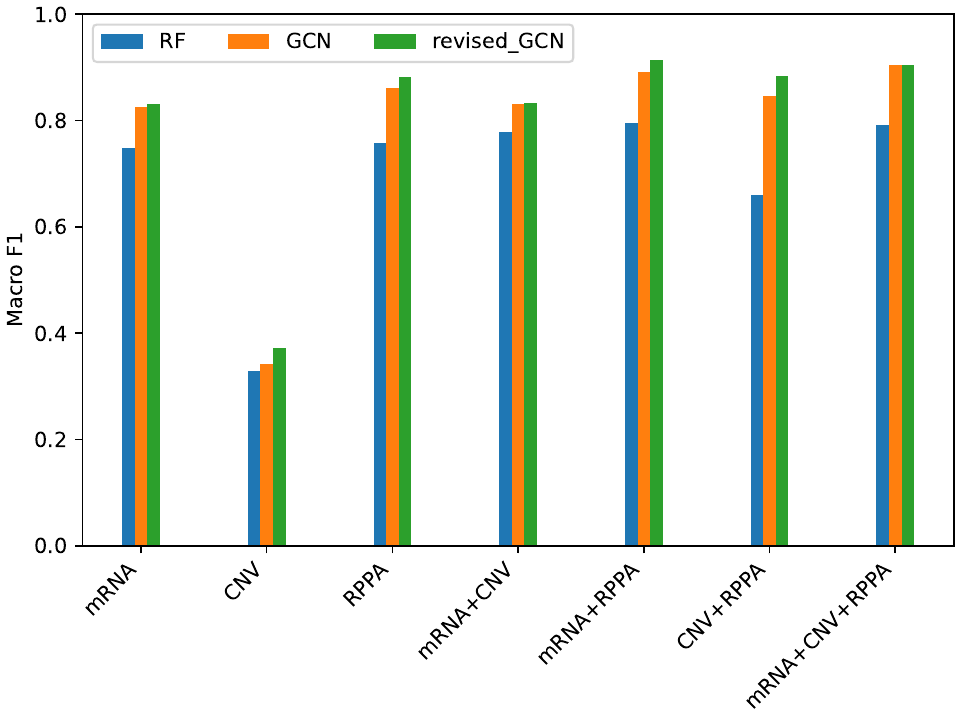}
\caption{Classification performance of different algorithms across various modalities.}
\label{fig_compareF1}
\end{figure}

To address this issue, we propose a revised GCN algorithm wherein the similarity network is consistently the fusion similarity network, irrespective of the input features. This adjustment ensures a more robust representation of sample relationships. As evidenced by the results in Table \ref{tab_revisedGCN} and Figure \ref{fig_compareF1}, this revised GCN outperforms both the original GCN and the random forest algorithm. Nevertheless, it is important to note that across all algorithms, the inclusion of CNV as a modality in the combined input consistently leads to a decline in model performance.

\begin{table}
\centering \caption{Classification performance across various modalities using the graph convolution network.}
\begin{tabular}{lccc}
\hline
 & Accuracy & AUC & Macro F1 \\ \hline
\makecell[l]{Node: mRNA \\ Edge: mRNA} & 0.8727 & 0.9528 & 0.8248  \\ \hline
\makecell[l]{Node: CNV \\ Edge: CNV} & 0.6066 & 0.7172 & 0.3428 \\ \hline
\makecell[l]{Node: RPPA \\ Edge: RPPA} & 0.8865 & 0.9658  & 0.8607  \\ \hline
\makecell[l]{Node: mRNA+CNV \\ Edge: mRNA+CNV} & 0.8767 & 0.9593 & 0.8318 \\ \hline
\makecell[l]{Node: mRNA+RPPA \\ Edge: mRNA+RPPA} & 0.9061 & 0.9838  & 0.8905 \\ \hline
\makecell[l]{Node: CNV+RPPA \\ Edge: CNV+RPPA} & 0.8727 & 0.9623  & 0.8464 \\ \hline
\makecell[l]{Node: mRNA+CNV+RPPA \\ Edge: mRNA+CNV+RPPA} & 0.9178 & 0.9840 & 0.9051 \\ \hline
\end{tabular}
\label{tab_GCN}
\end{table}

\begin{table}
\centering \caption{Classification performance across various modalities using the revised graph convolution network.}
\begin{tabular}{lccc}
\hline
 & Accuracy & AUC & Macro F1 \\ \hline
\makecell[l]{Node: mRNA \\ Edge: mRNA+CNV+RPPA} & 0.8747 & 0.9592  & 0.8314  \\ \hline
\makecell[l]{Node: CNV \\ Edge: mRNA+CNV+RPPA} & 0.6438  & 0.7836  & 0.3730  \\ \hline
\makecell[l]{Node: RPPA \\ Edge: mRNA+CNV+RPPA} & 0.9001 & 0.9773 & 0.8820  \\ \hline
\makecell[l]{Node: mRNA+CNV \\ Edge: mRNA+CNV+RPPA}  & 0.8786 & 0.9645 &  0.8328 \\ \hline
\makecell[l]{Node: mRNA+RPPA \\ Edge: mRNA+CNV+RPPA} & 0.9178  & 0.9848 & 0.9143  \\ \hline
\makecell[l]{Node: CNV+RPPA \\ Edge: mRNA+CNV+RPPA} & 0.9021  & 0.9768  & 0.8840  \\ \hline
\end{tabular}
\label{tab_revisedGCN}
\end{table}

\section{Conclusion}
We presented a balanced multimodal learning framework that unifies (1) an r-GCN encoder leveraging cross-modal similarity networks, (2) cross-modal self-distillation to elevate weak modalities, and (3) a multitask-like training scheme that reweights losses using macro-F1 and mutual information. On multi-omics BRCA classification, our framework alleviates dominance by strong modalities, suppresses the influence of low-information channels, and yields consistent performance gains over standard concatenation and unimodal models. Beyond the specific application, our results support a general recipe for multimodal learning under data scarcity and heterogeneity: \textbf{transfer what is reliable, quantify what is shared, and balance what is optimized}. Future work will extend our framework to settings with systematically missing modalities, explore alternative dependence measures beyond MI (e.g., Wasserstein-based criteria), and integrate uncertainty-aware task weighting to further stabilize training in ultra-low-sample cohorts.

\bibliographystyle{apalike} 
\bibliography{references} %

\end{document}